\documentclass[runningheads]{llncs}
\usepackage{graphicx}
\usepackage[ruled]{algorithm2e}
\usepackage{amsmath,amssymb} 
\usepackage{color}
\usepackage[width=122mm,left=12mm,paperwidth=146mm,height=193mm,top=12mm,paperheight=217mm]{geometry}
\usepackage{hyperref}

\begin{document}
\title{Learning to Navigate for Fine-grained Classification} 

\titlerunning{Learning to Navigate for Fine-grained Classification}
%
\author{Ze Yang\inst{1} \and
Tiange Luo\inst{1} \and
Dong Wang\inst{1} \and
Zhiqiang Hu\inst{1} \and
Jun Gao\inst{1} \and
Liwei Wang\inst{1,2}}
%
\authorrunning{Yang et al.}
%

\institute{Key Laboratory of Machine Perception, MOE, School of EECS, Peking University. \and Center for Data Science, Peking University, Beijing Institute of Big Data Research. \\\email{\{yangze,luotg,wangdongcis,huzq,jun.gao\}@pku.edu.cn\\wanglw@cis.pku.edu.cn}}

\maketitle              
\begin{abstract}
Fine-grained classification is challenging due to the difficulty of finding discriminative features. Finding those subtle traits that fully characterize the object is not straightforward. To handle this circumstance, we propose a novel self-supervision mechanism to effectively localize informative regions without the need of bounding-box/part annotations.  Our model, termed NTS-Net for Navigator-Teacher-Scrutinizer Network, consists of a Navigator agent, a Teacher agent and a Scrutinizer agent. In consideration of intrinsic consistency between informativeness of the regions and their probability being ground-truth class, we design a novel training paradigm,  which enables Navigator to detect most informative regions under the guidance from Teacher. After that, the Scrutinizer scrutinizes the proposed regions from Navigator and makes predictions. Our model can be viewed as a multi-agent cooperation, wherein agents benefit from each other, and make progress together. NTS-Net can be trained end-to-end, while provides accurate fine-grained classification predictions as well as highly informative regions during inference. We achieve state-of-the-art performance in extensive benchmark datasets.
\end{abstract}
\section{Introduction}\label{intro}

Fine-grained classification aims at differentiating subordinate classes of a common superior class, \emph{e.g.} distinguishing wild bird species, automobile models, \emph{etc}.  Those subordinate classes are usually defined by domain experts with complicated rules, which typically focus on subtle differences in particular regions. While deep learning has promoted the research in many computer vision~\cite{Krizhevsky2012ImageNet,ren2015faster,long_shelhamer_fcn} tasks, its application in fine-grained classification is more or less unsatisfactory, due in large part to the difficulty of finding informative regions and extracting discriminative features therein.  The situation is even worse for subordinate classes with varied poses like birds.

\begin{figure}[ht]
\begin{center}
\includegraphics[width=0.8\linewidth]{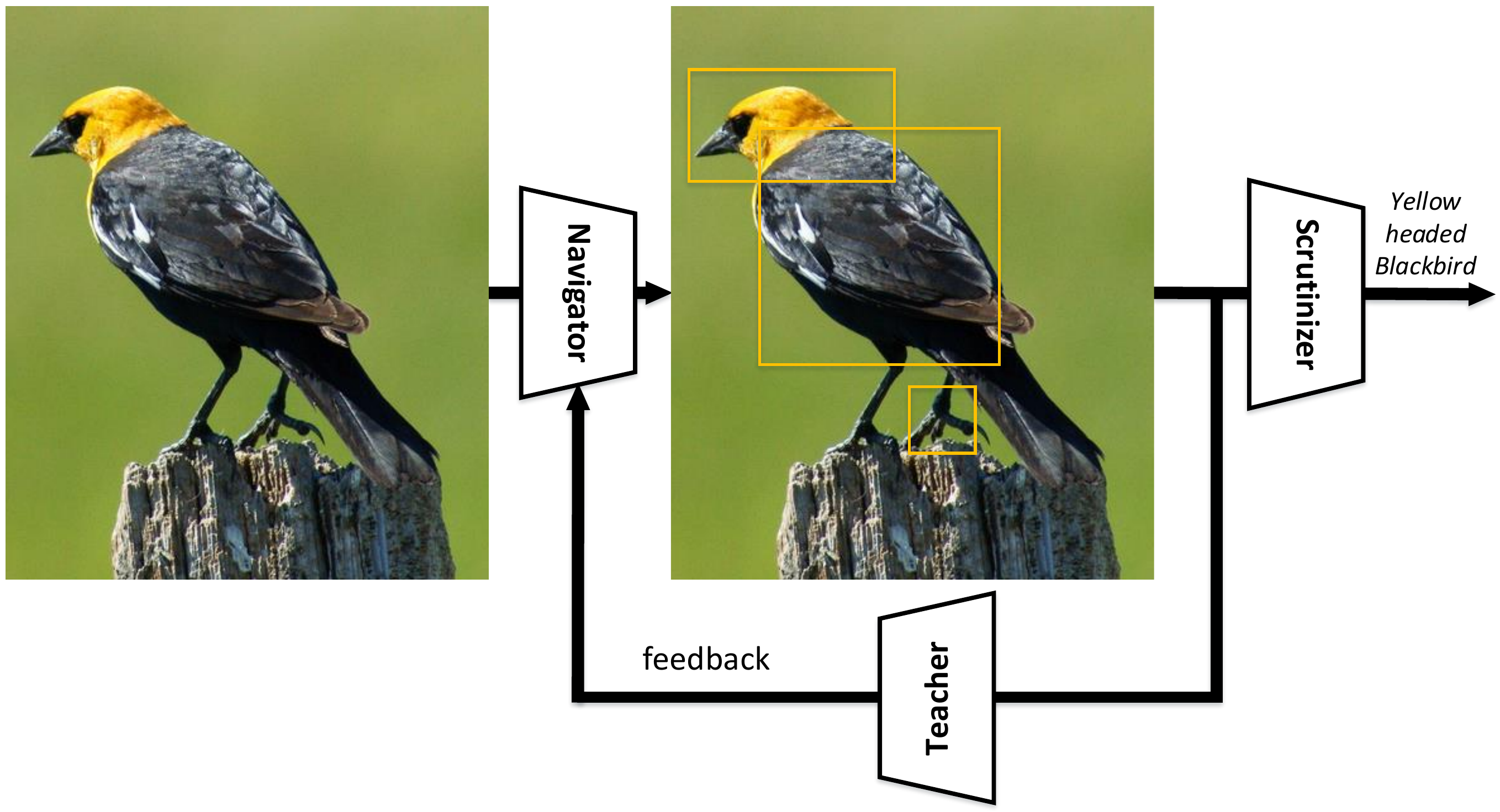}
\end{center}
   \caption{The overview of our model.  The Navigator navigates the model to focus on the most informative regions (denoted by yellow rectangles), while Teacher evaluates the regions proposed by Navigator and provides feedback. After that, the Scrutinizer scrutinizes those regions to make predictions.}
\label{overview}
\end{figure}

As a result, the key to fine-grained classification lies in developing automatic methods to accurately identify informative regions in an image. Some previous works~\cite{Xie2013Hierarchical,Chai2013Symbiotic,BransonVBP14,ZhangECCV14,Gavves2014Fine,Berg:2013:PPO:2514950.2516271,Liu2012Dog} take advantage of fine-grained human annotations, like annotations for bird parts in bird classification.  While achieving decent results, the fine-grained human annotations they require are expensive, making those methods less applicable in practice. Other methods~\cite{Zheng_2017_ICCV,Zhang_2016_CVPR,Zhao:2017:DVA:3101467.3101537,Wang2015Multiple} employ an unsupervised learning scheme to localize informative regions. They eliminate the need for the expensive annotations, but lack a mechanism to guarantee that the model focuses on the right regions, which usually results in degraded accuracy.

In this paper, we propose a novel self-supervised mechanism to effectively localize informative regions without the need of fine-grained bounding-box/part annotations.  The model we develop, which we term NTS-Net for Navigator-Teacher-Scrutinizer Network, employs a multi-agent cooperative learning scheme to address the problem of accurately identifying informative regions in an image.  Intuitively, the regions assigned higher probability to be ground-truth class should contain more object-characteristic semantics enhancing the classification performance of the whole image. Thus we design a novel loss function to optimize the informativeness of each selected region to have the same order as its probability being ground-truth class, and we take the ground-truth class of full image as the ground-truth class of regions. 

Specifically, our NTS-Net consists of a Navigator agent, a Teacher agent and a Scrutinizer agent. The Navigator navigates the model to focus on the most informative regions: for each region in the image, Navigator predicts how informative the region is, and the predictions are used to propose the most informative regions. The Teacher evaluates the regions proposed by Navigator and provides feedbacks: for each proposed region, the Teacher evaluates its probability belonging to ground-truth class; the confidence evaluations guide the Navigator to propose more informative regions with our novel ordering-consistent loss function. The Scrutinizer scrutinizes proposed regions from Navigator and makes fine-grained classifications: each proposed region is enlarged to the same size and the Scrutinizer extracts features therein; the features of regions and of the whole image are jointly processed to make fine-grained classifications. As a whole, our method can be viewed as an actor-critic \cite{Konda2002Actor} scheme in reinforcement learning, where the Navigator is the actor and the Teacher is the critic. With a more precise supervision provided by the Teacher, the Navigator will localize more informative regions, which in turn will benefit the Teacher. As a result, agents make progress together and end up with a model which provides accurate fine-grained classification predictions as well as highly informative regions. Fig.~\ref{overview} shows an overview of our methods.

Our main contributions can be summarized as follows:  
\begin{itemize}
\item
We propose a novel multi-agent cooperative learning scheme to address the problem of accurately identifying informative regions in the fine-grained classification task without bounding-box/part annotations.
\item
We design a novel loss function, which enables Teacher to guide Navigator to localize the most informative regions in an image by enforcing the consistency between regions' informativeness and their probability being ground-truth class.

\item
Our model can be trained end-to-end, while provides accurate fine-grained classification predictions as well as highly informative regions during inference.  We achieve state-of-the-art performance in extensive benchmark datasets.
\end{itemize}

The remainder of this paper is organized as follows: We will review the related work in Section.~\ref{related_work}. In Section.~\ref{methods} we will elaborate our methods.  Experimental results are presented and analyzed in Section.~\ref{experiments} and finally, Section.~\ref{conclusion} concludes.


\section{Related Work}\label{related_work}

\subsection{Fine-grained classification} 
There have been a variety of methods designed to distinguish fine-grained categories. Since some fine-grained classification datasets provide bounding-box/part annotations, early works~\cite{Xie2013Hierarchical,Chai2013Symbiotic,Berg:2013:PPO:2514950.2516271} take advantage of those annotations at both training and inference phase. However in practice when the model is deployed, no human annotations will be available. Later on, some works~\cite{BransonVBP14,ZhangECCV14} use bounding-box/part annotations only at training phase.  Under this setting, the framework is quite similar to detection: selecting regions and then classifying the pose-normalized objects. Besides, Jonathan \emph{et al.}~\cite{DengKrauseFei-Fei_CVPR2013} use co-segmentation and alignment to generate parts without part annotations but the bounding-box annotations are used during training.  Recently, a more general setting has emerged that does not require bounding box/part annotations either at training or inference time.  This setting makes fine-grained classification more useful in practice. This paper will mainly consider the last setting, where bounding-box/part annotations are not needed either at training or inference phase. 

In order to learn without fine-grained annotations, Jaderberg \emph{et al.}~\cite{NIPS2015_5854} propose Spatial Transformer Network to explicitly manipulate data representation within the network and predict the location of informative regions. Lin \emph{et al.}~\cite{lin2015bilinear} use a bilinear model to build discriminative features of the whole image; the model is able to capture subtle differences between different subordinate classes.  Zhang \emph{et al.}~\cite{Zhang_2016_CVPR} propose a two-step approach to learn a bunch of part detectors and part saliency maps. Fu \emph{et al.}~\cite{Fu_2017_CVPR} use an alternate optimization scheme to train attention proposal network and region-based classifier; they show that two tasks are correlated and can benefit each other. Zhao \emph{et al.}~\cite{Zhao:2017:DVA:3101467.3101537} propose Diversified Visual Attention Network (DVAN) to explicitly pursues the diversity of attention and better gather discriminative information. Lam \emph{et al.}~\cite{Lam_2017_CVPR} propose a Heuristic-Successor Network (HSNet) to formulate the fine-grained classification problem as a sequential search for informative regions in an image.

\subsection{Object detection}
Early object detection methods employ SIFT~\cite{Lowe2004Distinctive} or HOG~\cite{Dalal2005Histograms} features.  Recent works are mainly focusing on convolutional neural networks. Approaches like R-CNN~\cite{girshick2014rich}, OverFeat~\cite{Sermanet2013OverFeat} and SPPnet~\cite{He2015Spatial} adopt traditional image-processing methods to generate object proposals and perform category classification and bounding box regression.  Later works like Faster R-CNN~\cite{ren2015faster} propose Region Proposal Network (RPN) for proposal generation. YOLO~\cite{Redmon2016You} and SSD~\cite{Liu2016SSD} improve detection speed over Faster R-CNN~\cite{ren2015faster} by employing a single-shot architecture. On the other hand, Feature Pyramid Networks (FPN)~\cite{Lin_2017_CVPR} focuses on better addressing multi-scale problem and generates anchors from multiple feature maps. Our method requires selecting informative regions, which can also be viewed as object detection. To the best of our knowledge, we are the first one to introduce FPN into fine-grained classification while eliminates the need of human annotations.

\subsection{Learning to rank} 
Learning to rank is drawing attention in the field of machine learning and information retrieval~\cite{Liu:2009:LRI:1618303.1618304}. The training data consist of lists of items with assigned orders, while the objective is to learn the order for item lists. The ranking loss function is designed to penalize pairs with wrong order. Let $X = \{X_1, X_2, \cdots, X_n\}$ denote the objects to rank, and $Y = \{Y_1, Y_2, \cdots, Y_n\}$ the indexing of the objects,  where $Y_i \geq Y_j$ means $X_i$ should be ranked before $X_j$. Let $\mathbb F$ be the hypothesis set of ranking function.  The goal is to find a ranking function $\mathcal F \in \mathbb F$ that minimize a certain loss function defined on $\{X_1, X_2 \cdots X_n\}$, $\{Y_1, Y_2, \cdots, Y_n\}$ and $\mathcal F$. There are many ranking methods. Generally speaking, these methods can be divided into three categories: the point-wise approach~\cite{Cossock2008Statistical}, pair-wise approach~\cite{Herbrich2000Large,Burges2005Learning} and list-wise approach\cite{Cao2007Learning,Xia2008Listwise}.

Point-wise approach assign each data with a numerical score, and the learning-to-rank problem can be formulated as a regression problem,  for example with $L2$ loss function:
\begin{equation}
	\label{point-wise}
	L_{point}(\mathcal F, X, Y) = \sum_{i=1}^n (\mathcal F(X_i)-Y_i)^2 
\end{equation}

In the pair-wise ranking approach, the learning-to-rank problem is formulated as a classification problem. \emph{i.e.} to learn a binary classifier that chooses the superiority in a pair. Suppose $\mathcal F(X_i,X_j)$ only takes a value from $\{1, 0\}$, where $\mathcal F(X_i,X_j) = 0$ means $X_i$ is ranked before $X_j$. Then the loss is defined on all pairs as in Eqn.~\ref{pair-wise}, and the goal is to find an optimal $\mathcal F$ to minimize the average number of pairs with wrong order.

\begin{equation}
	\label{pair-wise}
	L_{pair}(\mathcal F, X, Y) = \sum_{(i,j):Y_i < Y_j} \mathcal F(X_i, X_j)
\end{equation}

List-wise approach directly optimizes the whole list, and it can be formalized as a classification problem on permutations. Let $\mathcal F(X, Y)$ be the ranking function, the loss is defined as:

\begin{equation}
	\label{list-wise}
	L_{list}(\mathcal F, X, Y)= 
	\begin{cases}
    	1,   & \text{if  }\mathcal F(X) \neq Y \\
    	0,   & \text{if  }\mathcal F(X) = Y
	\end{cases}
\end{equation}

In our approach, our navigator loss function adopts from the multi-rating pair-wise ranking loss, which enforces the consistency between region's informativeness and probability being ground-truth class. 

\section{Methods}\label{methods}
\subsection{Approach Overview}
Our approach rests on the assumption that informative regions are helpful to better characterize the object, so fusing features from informative regions and the full image will achieve better performance. Therefore the goal is to localize the most informative regions of the objects. We assume all regions\footnote{Without loss of generality, we also treat full image as a region} are rectangle, and we denote $\mathbb A$ as the set of all regions in the given image\footnote{Notation: we use $\mathcal C$alligraphy font to denote mapping, $\mathbb B$lackboard bold font to denote special sets, And we use $\mathbf B$old font to denote parameters in network.}. We define information function $\mathcal I: \mathbb A \to (-\infty, \infty)$ evaluating how informative the region $R \in \mathbb A$ is, and we define the confidence function $\mathcal C: \mathbb A \to [0,1]$ as a classifier to evaluate the confidence that the region belongs to ground-truth class. As mentioned in Sec.~\ref{intro}, more informative regions should have higher confidence, so the following condition should hold:

\begin{itemize}
 	\item[$\bullet$] Condition.~1: for any $ R_1, R_2 \in \mathbb A$, if $\mathcal C(R_1) > \mathcal C(R_2)$, $\mathcal I(R_1) > \mathcal I(R_2)$
\end{itemize}

We use Navigator network to approximate information function $\mathcal I$ and Teacher network to approximate confidence function $\mathcal C$. For the sake of simplicity, we choose $M$ regions $\mathbb A_M$ in the region space $\mathbb A$. For each region $R_i \in \mathbb A_M$, the Navigator network evaluates its informativeness $\mathcal I(R_i)$, and the Teacher network evaluates its confidence
 $\mathcal C(R_i)$. In order to satisfy Condition.~1, we optimize Navigator network to make $\{\mathcal I(R_1), \mathcal I(R_2), \cdots, \mathcal I(R_M)\}$ and $\{\mathcal C(R_1), \mathcal C(R_2), \cdots, \mathcal C(R_M)\}$ having the same order. 

As the Navigator network improves in accordance with the Teacher network, it will produce more informative regions to help Scrutinizer network make better fine-grained classification result.

In Section.~\ref{sec_navigator_teacher}, we will describe how informative regions are proposed by Navigator under Teacher's supervision. In Section.~\ref{sec_scrutinizer},  we will present how to get fine-grained classification result from Scrutinizer. In Section.~\ref{sec_architecture} and ~\ref{sec_loss}, we will introduce the network architecture and optimization in detail, respectively.

\subsection{Navigator and Teacher}\label{sec_navigator_teacher}
\begin{figure}[ht]
\begin{center}
\includegraphics[width=0.75\linewidth]{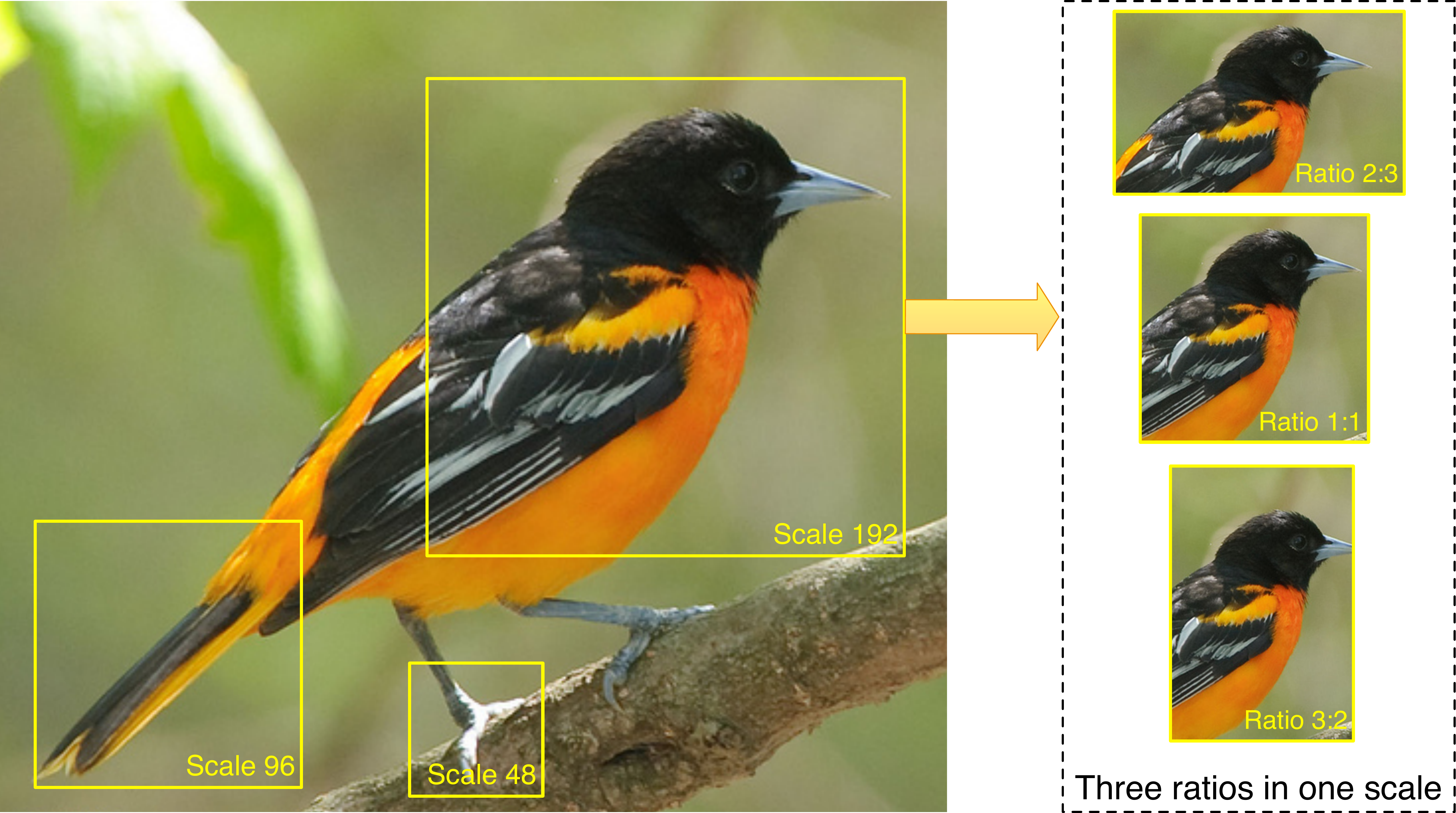}
\end{center}
   \caption{The design of anchors. We use three scales and three ratios. For an image of size 448, we construct anchors to have scales of \{48, 96, 192\} and ratios \{1:1, 2:3, 3:2\}.}
\label{anchors}
\end{figure}

Navigating to possible informative regions can be viewed as a region proposal problem, which has been widely studied in~\cite{Uijlings2013Selective,Endres2010Category,Arbelaez2014Multiscale,Carreira2012CPMC,NIPS2016_6532}. Most of them are based on a sliding-windows search mechanism. Ren \emph{et al.}~\cite{ren2015faster} introduce a novel region proposal network (RPN) that shares convolutional layers with the classifier and mitigates the marginal cost for computing proposals. They use anchors to simultaneously predict multiple region proposals. Each anchor is associated with a sliding window position, aspect ratio, and box scale. Inspired by the idea of anchors, our Navigator network takes an image as input, and produce a bunch of rectangle regions $\{R'_1, R'_2, \dots R'_A\}$, each with a score denoting the informativeness of the region (Fig.~\ref{anchors} shows the design of our anchors). For an input image $X$ of size 448, we choose anchors to have scales of \{48, 96, 192\} and ratios \{1:1, 3:2, 2:3\}, then Navigator network will produce a list denoting the informativeness of all anchors. We sort the information list as in Eqn.~\ref{sort} , where $A$ is the number of anchors, $\mathcal I(R_i)$ is the $i$-th element in sorted information list.

\begin{equation}\label{sort}
	\mathcal I(R_1) \geq \mathcal I(R_2) \geq \dots \geq \mathcal I(R_A)
\end{equation} 

\begin{figure}[ht]
\begin{center}
\includegraphics[width=0.85\linewidth]{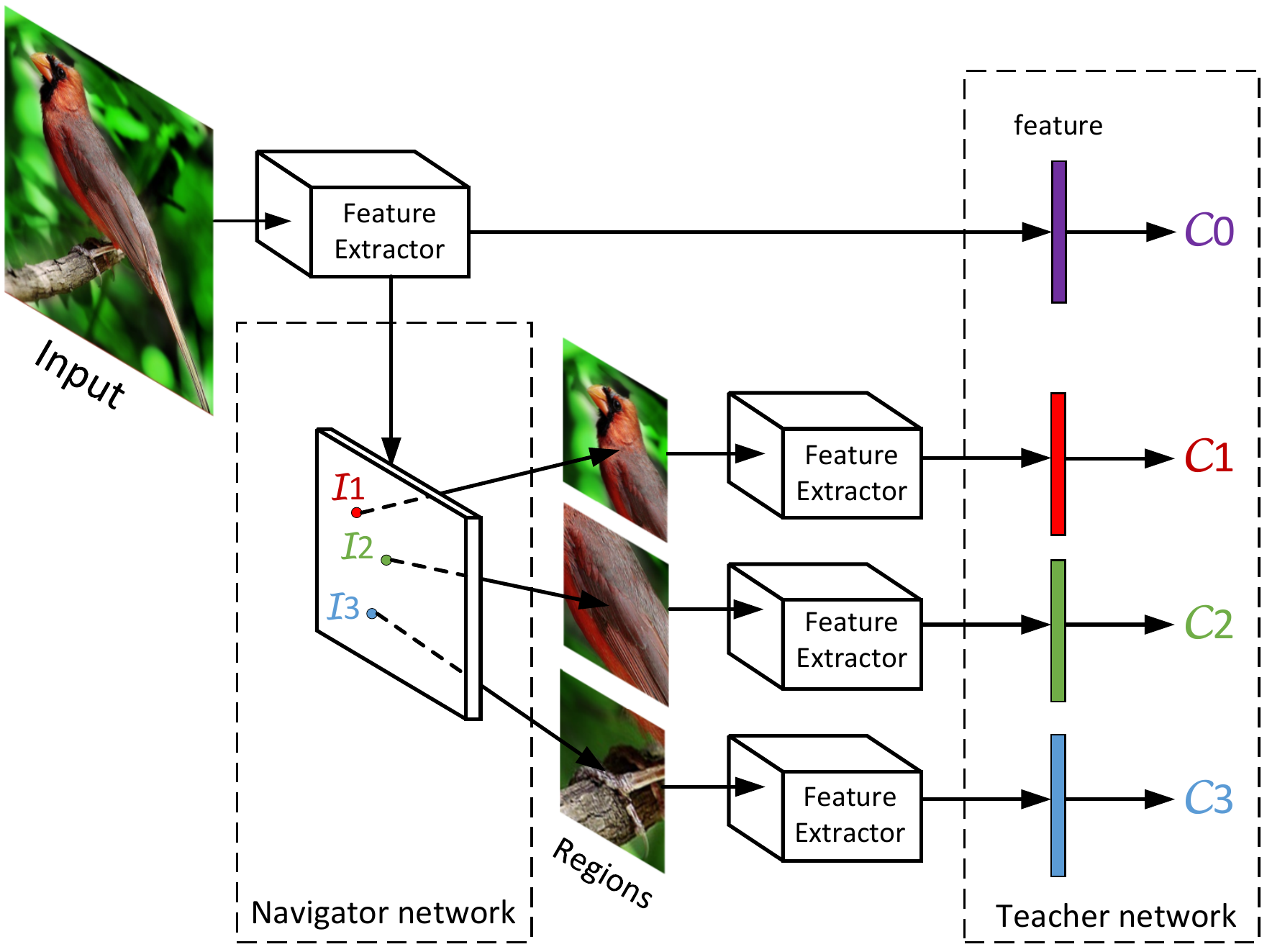}
\end{center}
   \caption{Training method of Navigator network. For an input image, the feature extractor extracts its deep feature map, then the feature map is fed into Navigator network to compute the informativeness of all regions. We choose top-$M$ (here $M=3$ for explanation) informative regions after NMS and denote their informativeness as $\{I_1, I_2, I_3\}$. Then we crop the regions from the full image, resize them to the pre-defined size and feed them into Teacher network, then we get the confidences $\{C_1, C_2, C_3\}$. We optimize Navigator network to make $\{I_1, I_2, I_3\}$ and $\{C_1, C_2, C_3\}$ having the same order.}
\label{teacher}
\end{figure}

To reduce region redundancy, we adopt non-maximum suppression (NMS) on the regions based on their informativeness. Then we take the top-$M$ informative regions $\{R_1, R_2, \dots, R_M\}$ and feed them into the Teacher network to get the confidence  as $\{\mathcal C(R_1), \mathcal C(R_2), \dots \mathcal C(R_M)\}$. Fig.~\ref{teacher} shows the overview with $M = 3$, where $M$ is a hyper-parameters denoting how many regions are used to train Navigator network. We optimize Navigator network to make $\{\mathcal I(R_1), \mathcal I(R_2), \dots \mathcal I(R_M)\}$ and $\{\mathcal C(R_1), \mathcal C(R_2), \dots \mathcal C(R_M)\}$ having the same order. Every proposed region is used to optimize Teacher by minimizing the cross-entropy loss between ground-truth class and the predicted confidence.

\subsection{Scrutinizer}\label{sec_scrutinizer}
\begin{figure}[ht]
\begin{center}
\includegraphics[width=0.85\linewidth]{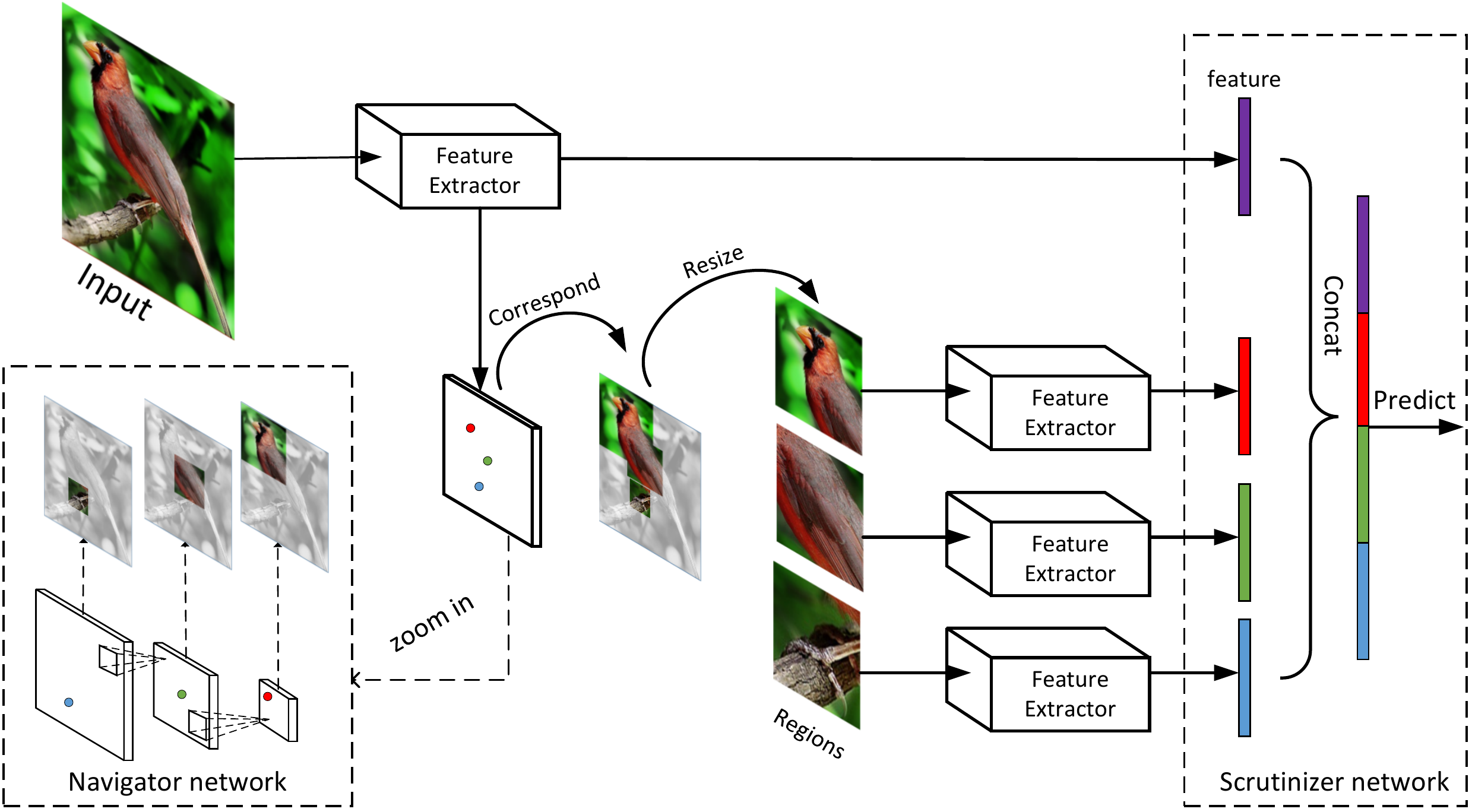}
\end{center}
   \caption{Inference process of our model (here $K=3$ for explanation). The input image is first fed into feature extractor, then the Navigator network proposes the most informative regions of the input. We crop these regions from the input image and resize them to the pre-defined size, then we use feature extractor to compute the features of these regions and fuse them with the feature of the input image. Finally, the Scrutinizer network processes the fused feature to predict labels.}
\label{scrutinizer}
\end{figure}

As Navigator network gradually converges, it will produce informative object-characteristic regions to help Scrutinizer network make decisions. We use the top-$K$ informative regions combined with the full image as input to train the Scrutinizer network. In other words,  those $K$ regions are used to facilitate fine-grained recognition. Fig.~\ref{scrutinizer} demonstrates this process with $K=3$. Lam \emph{et al.}~\cite{Lam_2017_CVPR} show that using informative regions can reduce intra-class variance and are likely to generate higher confidence scores on the correct label. Our comparative experiments show that adding informative regions substantially improve fine-grained classification results in a wide range of datasets including CUB-200-2001, FGVC Aircraft, and Stanford Cars, which are shown in Table.~\ref{birds},~\ref{aircraft_car}.

\subsection{Network architecture}\label{sec_architecture}
In order to obtain correspondence between region proposals and feature vectors in feature map, we use fully-convolutional network as the feature extractor, without fully-connected layers. Specifically, we choose ResNet-50~\cite{he2016deep} pre-trained on ILSVRC2012~\cite{ILSVRC15} as the CNN feature extractor, and Navigator, Scrutinizer, Teacher network all share parameters in feature extractor. We denote parameters in feature extractor as $\mathbf W $. For input image $X$, the extracted deep representations are denoted as $X \otimes \mathbf W$, where $\otimes$ denotes the combinations of convolution, pooling, and activation operations.\\\\
\textbf{Navigator network.}
Inspired by the design of Feature Pyramid Networks (FPN)~\cite{Lin_2017_CVPR}, we use a top-down architecture with lateral connections to detect multi-scale regions. We use convolutional layers to compute feature hierarchy layer by layer, followed by ReLU activation and max-pooling. Then we get a series of feature maps of different spatial resolutions. The anchors in larger feature maps correspond to smaller regions. Navigator network in Figure.~\ref{scrutinizer} shows the sketch of our design. Using multi-scale feature maps from different layers we can generate informativeness of regions among different scales and ratios. In our setting, we use feature maps of size $\{14 \times 14, 7 \times 7, 4 \times 4\}$ corresponding to regions of  scale $\{48 \times 48, 96 \times 96, 192 \times 192\}$. We denote the parameters in Navigator network as $\mathbf W_{\mathcal I}$ (including shared parameters in feature extractor).\\\\
\textbf{Teacher network.}
The Teacher network (Fig.~\ref{teacher}) approximates the mapping $\mathcal C: \mathbb A \to [0,1]$ which denotes the confidence of each region. After receiving $M$ scale-normalized ($224 \times 224$) informative regions $\{R_1, R_2, \dots, R_M\}$ from Navigator network, Teacher network outputs confidence as teaching signals to help Navigator network learn. In addition to the shared layers in feature extractor, the Teaching network has a fully connected layer which has $2048$ neurons. We denote the parameters in Teacher network as $\mathbf W_{\mathcal C}$ for convenience.\\\\
\textbf{Scrutinizer network.}
After receiving top-$K$ informative regions from Navigator network, the $K$ regions are resized to the pre-defined size (in our experiments we use $224 \times 224$) and are fed into feature extractor to generate those $K$ regions' feature vector, each with length $2048$. Then we concatenate those $K$ features with input image's feature, and feed it into a fully-connected layer which has $2048\times(K+1)$ neurons (Fig.~\ref{scrutinizer}).  We use function $\mathcal S$ to represent the composition of these transformations. We denote the parameters in Scrutinizer network as $\mathbf W_{\mathcal S}$. 

\subsection{Loss function and Optimization}\label{sec_loss}
\textbf{Navigation loss.}
We denote the $M$ most informative regions predicted by Navigator network as $R = \{R_1, R_2, \dots, R_M\}$, their informativeness as $I = \{I_1, I_2, \dots, I_M\}$, and their confidence predicted by Teacher network as $C = \{C_1, C_2, \dots, C_M\}$. Then the navigation loss is defined as follow:
\begin{equation}\label{navigator_loss}
	L_{\mathcal I}(I, C) = \sum_{(i,s):C_i < C_s} f(I_s-I_i)
\end{equation}
where the function $f$ is a non-increasing function that encourages $I_s>I_i$ if $C_s>C_i$, and we use hinge loss function $f(x) = \max\{1-x,0\}$ in our experiment. The loss function penalize reversed pairs\footnote{Given a list $x=\{x_1, x_2, \cdots, x_n\}$ be the data and a permutation $\pi=\{\pi_1, \pi_2, \cdots, \pi_n\}$ be the order of the data. Reverse pairs are pairs of elements in $x$ with reverse order. \emph{i.e.} if $x_i<x_j$ and $\pi_i>\pi_j$ holds at same time, then $x_i$ and $x_j$ is an reverse pair.} between $I$ and $C$, and encourage that $I$ and $C$ is in the same order. Navigation loss function is differentiable, and calculating the derivative \emph{w.r.t.} $\mathbf W_{\mathcal I}$ by the chain rule in back-propagation we get:
\begin{align}\label{navigator_loss_doff}
	&\frac{\partial L_{\mathcal I}(I, C)}{\partial \mathbf W_{\mathcal I}} \\
	= &\sum_{(i,s):C_i < C_s} f'(I_s-I_i)\cdot (\frac{\partial \mathcal I(x)}{\partial \mathbf W_{\mathcal I}}\Bigr |_{x = R_s}-\frac{\partial \mathcal I(x)}{\partial \mathbf W_{\mathcal I}}\Bigr |_{x = R_i}) \nonumber
\end{align}

The equation follows directly by the definition of $I_i = \mathcal I(R_i)$.\\\\
\textbf{Teaching loss.}
We define the Teacher loss $L_{\mathcal C}$ as follows:

\begin{equation}\label{teachloss}
	L_{\mathcal C} = -\sum_{i=1}^M\log \mathcal C(R_i) - \log \mathcal C(X)
\end{equation}
where $\mathcal C$ is the confidence function which maps the region to its probability being ground-truth class. The first term in Eqn. \ref{teachloss} is the sum of cross entropy loss of all regions, the second term is the cross entropy loss of full image.\footnote{The second term helps training. For simplicity, we also denote the confidence function of full image as $\mathcal C$. }\\\\
\textbf{Scrutinizing loss.}
When the Navigator network navigates to the most informative regions $\{R_1, R_2, \cdots, R_K\}$, the Scrutinizer network makes the fine-grained recognition result $P = \mathcal S(X, R_1, R_2, \cdots, R_K)$. We employ cross entropy loss as classification loss:
\begin{equation}
	L_{\mathcal S} = -\log \mathcal S(X, R_1, R_2, \cdots, R_K)
\end{equation}\\
\textbf{Joint training algorithm.} 
The total loss is defined as:
\begin{equation}\label{total_loss}
	L_{total} = L_{\mathcal I} + \lambda \cdot L_{\mathcal S} + \mu \cdot L_{\mathcal C}
\end{equation}
where $\lambda$ and $\mu$ are hyper-parameters. In our setting, $\lambda = \mu = 1$. The overall algorithm is summarized in Algorithm.~\ref{alg1}. We use stochastic gradient method to optimize $L_{total}$.

\begin{algorithm}
\caption{NTS-Net algorithm}
\label{alg1}
\LinesNumbered
\KwIn{full image $X$, hyper-parameters $K$, $M$, $\lambda$, $\mu$, assume $K \leq M$}
\KwOut{predict probability $P$}
\For{t = 1,T}{
	Take full image $=X$\\
	Generate anchors $\{R'_1,R'_2, \dots, R'_A\}$\\
	$\{I'_1, \dots, I'_A\}:=\mathcal I(\{R'_1, \dots, R'_A\})$\\
	$\{I_i\}_{i=1}^A, \{R_i\}_{i=1}^A := $ NMS($\{I'_i\}_{i=1}^A, \{R'_i\}_{i=1}^A$)\\
	Select top $M$: $\{I_i\}_{i=1}^M, \{R_i\}_{i=1}^M$\\
	$\{C_1, \dots, C_K\} := \mathcal C(\{R_1, \dots, R_K\})$\\
	$P = \mathcal S(X, R_1, R_2, \cdots, R_K)$\\
	Calculate $L_{total}$ from Eqn. \ref{total_loss}\\
	BP$(L_{total})$ get gradient \emph{w.r.t.} $\mathbf W_{\mathcal I}$, $\mathbf W_{\mathcal C}$, $\mathbf W_{\mathcal S}$\\
	Update $\mathbf W_{\mathcal I}$, $\mathbf W_{\mathcal C}$, $\mathbf W_{\mathcal S}$ using SGD\\
	}
\end{algorithm}

\section{Experiments}\label{experiments}
\subsection{Dataset}
We comprehensively evaluate our algorithm on Caltech-UCSD Birds (CUB-200-2011)~\cite{WahCUB_200_2011}, Stanford Cars~\cite{Krause20133D} and FGVC Aircraft~\cite{maji13fine-grained} datasets, which are widely used benchmark for fine-grained image classification. We do not use any bounding box/part annotations in all our experiments. Statistics of all 3 datasets are shown in Table.~\ref{data_stat}, and we follow the same train/test splits as in the table.
\textbf{Caltech-UCSD Birds.}
CUB-200-2011 is a bird classification task with 11,788 images from 200 wild bird species. The ratio of train data and test data is roughly $1:1$. It is generally considered one of the most competitive datasets since each species has only 30 images for training.\\
\textbf{Stanford Cars.}
Stanford Cars dataset contains 16,185 images over 196 classes, and each class has a roughly 50-50 split. The cars in the images are taken from many angles, and the classes are typically at the level of production year and model (\emph{e.g.} 2012 Tesla Model S).\\
\textbf{FGVC Aircraft.}
FGVC Aircraft dataset contains 10,000 images over 100 classes, and the train/test set split ratio is around $2:1$. Most images in this dataset are airplanes. And the dataset is organized in a four-level hierarchy, from finer to coarser: Model, Variant, Family, Manufacturer.

\begin{table}[ht]
	\begin{center}
	\begin{tabular}{|c|c|c|c|c|}
	\hline
	Dataset & $\#$Class & $\#$Train & $\#$Test \\
	\hline\hline
	CUB-200-2011 & $200$ & $5,994$ & $5,794$ \\
	\hline
	Stanford Cars & $196$ & $8,144$ & $8,041$ \\
	\hline
	FGVC Aircraft & $100$ & $6,667$ & $3,333$ \\
	\hline
	\end{tabular}
	\end{center}
\caption{Statistics of benchmark datasets.}
\label{data_stat}
\end{table}

\subsection{Implementation Details}
In all our experiments, we preprocess images to size $448 \times 448$, and we fix $M=6$ which means $6$ regions are used to train Navigator network for each image (there is no restriction on hyper-parameters $K$ and $M$). We use fully-convolutional network ResNet-50~\cite{he2016deep} as feature extractor and use Batch Normalization as regularizer. We use Momentum SGD with initial learning rate $0.001$ and multiplied by $0.1$ after $60$ epochs, and we use weight decay $1\mathrm{e}{-4}$. The NMS threshold is set to $0.25$, no pre-trained detection model is used. Our model is robust to the selection of hyper-parameters. We use Pytorch to implement our algorithm and the code will be available at \url{https://github.com/yangze0930/NTS-Net}.

\subsection{Quantitative Results}
Overall, our proposed system outperforms all previous methods. Since we do not use any bounding box/part annotations, we do not compare with methods which depend on those annotations. Table.~\ref{birds} shows the comparison between our results and previous best results in CUB-200-2011. ResNet-50 is a strong baseline, which by itself achieves $84.5\%$ accuracy, while our proposed NTS-Net outperforms it by a clear margin $3.0\%$. Compared to~\cite{Li_2017_ICCV_Workshops} which also use ResNet-50 as feature extractor, we achieve a $1.5\%$ improvement. It is worth noting that when we use only full image ($K=0$) as input to the Scrutinizer, we achieve $85.3\%$ accuracy, which is also higher than ResNet-50. This phenomenon demonstrates that, in navigating to informative regions, Navigator network also facilitates Scrutinizer by sharing feature extractor, which learns better feature representation.
\begin{table}[ht]
	\begin{center}
	\begin{tabular}{|c|c|}
	\hline
	Method & top-$1$ accuracy \\
	\hline\hline
	MG-CNN~\cite{Wang2015Multiple} & $81.7\%$ \\
	\hline
	Bilinear-CNN~\cite{lin2015bilinear} & $84.1\%$ \\
	\hline
	ST-CNN~\cite{NIPS2015_5854} & $84.1\%$ \\
	\hline
	FCAN~\cite{FCAN} & $84.3\%$\\
	\hline
	~~ResNet-50 (implemented in \cite{Li_2017_ICCV_Workshops})~~ & $84.5\%$ \\
	\hline
	PDFR~\cite{Zhang_2016_CVPR} & $84.5\%$\\
	\hline
	RA-CNN~\cite{Fu_2017_CVPR} & $85.3\%$\\
	\hline
	HIHCA~\cite{Cai_2017_ICCV} & $85.3\%$\\
	\hline
	Boost-CNN~\cite{Moghimi2016Boosted} & $85.6\%$\\
	\hline
	DT-RAM~\cite{Li_2017_ICCV_Workshops} & $86.0\%$\\
	\hline
	MA-CNN~\cite{Zheng_2017_ICCV} & $86.5\%$\\
	\hline
	\hline
	Our NTS-Net (K = 2) & $87.3\%$\\
	\hline
	Our NTS-Net (K = 4) & $\mathbf{87.5\%}$\\
	\hline
	\end{tabular}
	\end{center}
\caption{Experimental results in CUB-200-2011.}
\label{birds}
\end{table}

Table.~\ref{aircraft_car} shows our result in FGVC Aircraft and Stanford Cars, respectively. Our model achieves new state-of-the-art results with $91.4\%$ top-1 accuracy in FGVC Aircraft and $93.9\%$ top-1 accuracy in Stanford Cars.

\begin{table}[ht]
	\begin{center}
	\begin{tabular}{|c|c|c|}
	\hline
	Method & top-$1$ on FGVC Aircraft & top-$1$ on Stanford Cars \\
	\hline\hline
	FV-CNN~\cite{Gosselin2014Revisiting} & $81.5\%$ & - \\
	\hline
	FCAN~\cite{FCAN} & - & $89.1\%$ \\
	\hline
	Bilinear-CNN~\cite{lin2015bilinear} & $84.1\%$ & $91.3\%$ \\
	\hline
	RA-CNN~\cite{Fu_2017_CVPR} & $88.2\%$ & $92.5\%$\\
	\hline
	HIHCA~\cite{Cai_2017_ICCV} & $88.3\%$ & $91.7\%$\\
	\hline
	Boost-CNN~\cite{Moghimi2016Boosted} & $88.5\%$ & $92.1\%$\\
	\hline
	MA-CNN~\cite{Zheng_2017_ICCV} & $89.9\%$ & $92.8\%$\\
	\hline
	DT-RAM~\cite{Li_2017_ICCV_Workshops} & - & $93.1\%$\\
	\hline
	\hline
	~~~~~~ Our NTS-Net (K = 2) ~~~~~~ & $90.8\%$ & $93.7\%$\\
	\hline
	~~~~~~ Our NTS-Net (K = 4) ~~~~~~ & $\mathbf{91.4\%}$ & $\mathbf{93.9\%}$\\
	\hline
	\end{tabular}
	\end{center}
\caption{Experimental results in FGVC Aircraft and Stanford Cars.}
\label{aircraft_car}
\end{table}
\subsection{Ablation Study}
In order to analyze the influence of different components in our framework, we design different runs in CUB-200-2011 and report the results in Table.~\ref{factor}. We use NS-Net to denote the model without Teacher's guidance, NS-Net let the Navigator network alone to propose regions and the accuracy drops from $87.5\%$ to $83.3\%$, we hypothesize it is because the navigator receives no supervision from teacher and will propose random regions, which we believe cannot benefit classification. We also study the role of hyper-parameter $K$, \emph{i.e.} how many part regions have been used for classification. Referring to Table.~\ref{factor}, accuracy only increases $0.2\%$ when $K$ increases from 2 to 4, the accuracy improvement is minor while feature dimensionality nearly doubles. On the other hand, accuracy increases $2.0\%$ when $K$ increases from 0 to 2, which demonstrate simply increasing feature dimensionality will only get minor improvement, but our multi-agent framework will achieve considerable improvements ($0.2\%$ vs $2\%$). 

\begin{table}[ht]
	\begin{center}
	\begin{tabular}{|c|c|}
	\hline
	Method & top-$1$ accuracy \\
	\hline\hline
	ResNet-50 baseline & $84.5\%$ \\
	\hline
	NS-Net (K = 4) & $83.3\%$ \\
	\hline
	Our NTS-Net (K = 0) & $85.3\%$\\
	\hline
	Our NTS-Net (K = 2) & $87.3\%$\\
	\hline
	~~~~~~ Our NTS-Net (K = 4) ~~~~~~ & $\mathbf{87.5\%}$\\
	\hline
	\end{tabular}
	\end{center}
\caption{Study of influence factor in CUB-200-2011.}
\label{factor}
\end{table}
\subsection{Qualitative Results}
\begin{figure}[ht]
\begin{center}
\includegraphics[width=0.9\linewidth]{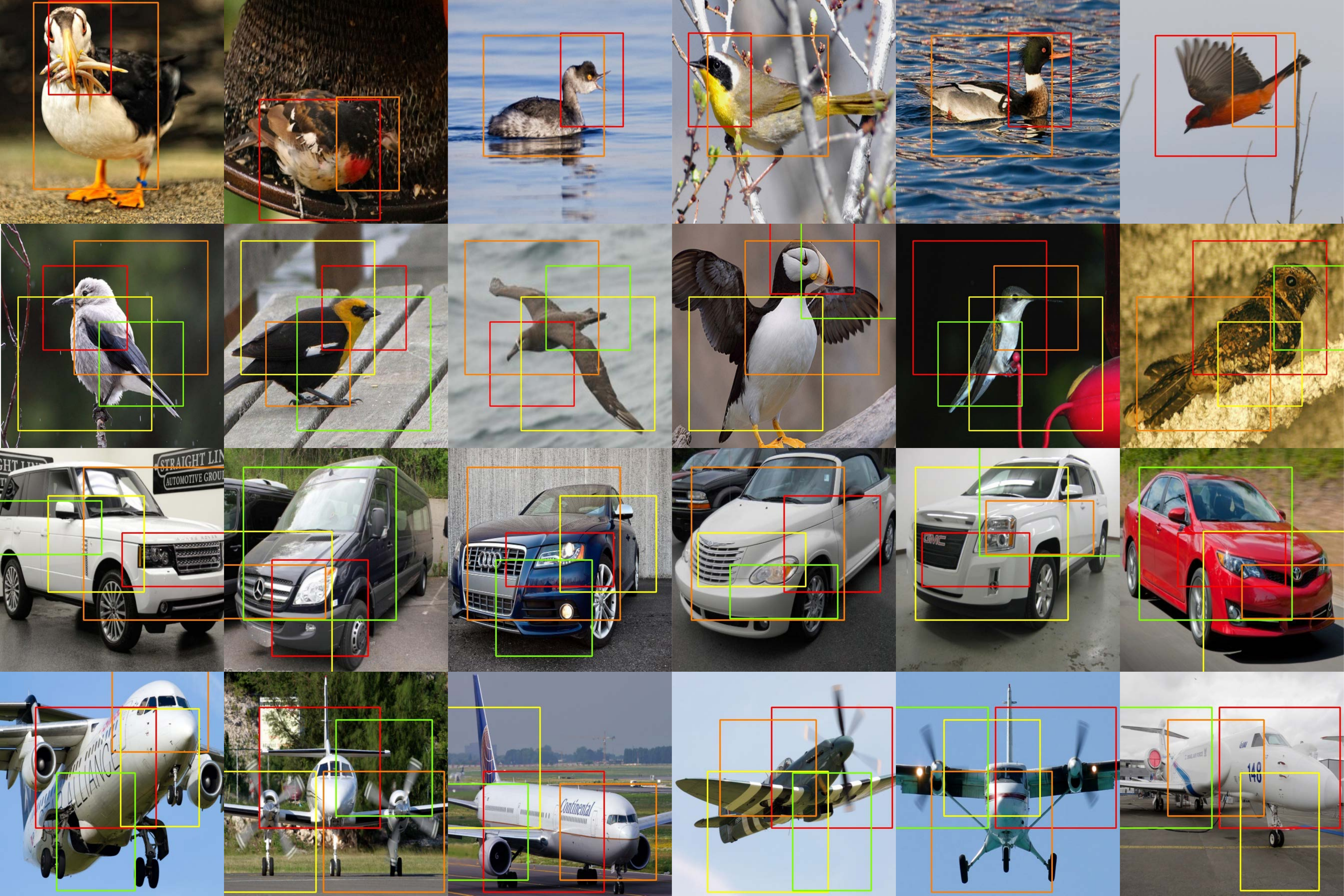}
\end{center}
   \caption{The most informative regions proposed by Navigator network. The first row shows $K=2$ in CUB-200-2011 dataset. The second to fourth rows show $K=4$ in CUB-200-2011, Stanford Cars and FGVC Aircraft, respectively.}
\label{quality}
\end{figure}

To analyze where Navigator network navigates the model, we draw the navigation regions predicted by Navigator network in Fig.~\ref{quality}. We use red, orange, yellow, green rectangles to denote the top four informative regions proposed by Navigator network, with red rectangle denoting most informative one. It can be seen that the localized regions are indeed informative for fine-grained classification. The first row shows $K=2$ in CUB-200-2011 dataset: we can find that using two regions are able to cover informative parts of birds, especially in the second picture where the color of the bird and the background is quite similar. The second row shows $K=4$ in CUB-200-2011: we can see that the most informative regions of birds are head, wings and main body, which is consistent with the human perception. The third row shows $K=4$ in Stanford Cars: we can find that the headlamps and grilles are considered the most informative regions of cars. The fourth row shows $K=4$ in FGVC Airplane: the Navigator network locates the airplane wings and head, which are very helpful for classification. 

\section{Conclusions}\label{conclusion}
In this paper, we propose a novel method for fine-grained classification without the need of bounding box/part annotations. The three networks, Navigator, Teacher and Scrutinizer cooperate and reinforce each other. We design a novel loss function considering the ordering consistency between regions' informativeness and probability being ground-truth class. Our algorithm is end-to-end trainable and achieves state-of-the-art results in CUB-200-2001, FGVC Aircraft and Stanford Cars datasets.

\section{Acknowledgments}
This work is supported by National Basic Research Program of China (973 Program) (grant no. 2015CB352502), NSFC (61573026) and BJNSF (L172037).


\end{document}